\RequirePackage{cmap}

\documentclass[conference,10pt]{IEEEtran}[2015/08/26]

\usepackage[zerostyle=b,scaled=.75]{newtxtt}

\usepackage{mwe}

\usepackage[T1]{fontenc}
\usepackage[utf8]{inputenc} 

\usepackage{graphicx}

\usepackage[ngerman,english]{babel}
\addto\extrasenglish{\languageshorthands{ngerman}\useshorthands{"}}

\usepackage{upquote}

\usepackage{paralist}

\usepackage{csquotes}

\usepackage{algorithm} 
\usepackage{algorithmic} 

\usepackage{courier}

\RequirePackage{iftex}
\ifPDFTeX
  \RequirePackage[%
    final,%
    expansion=alltext,%
    protrusion=alltext-nott]{microtype}%
\else
  \RequirePackage[%
    final,%
    protrusion=alltext-nott]{microtype}%
\fi%
\DisableLigatures{encoding = T1, family = tt* }

\usepackage{url}
\makeatletter
\g@addto@macro{\UrlBreaks}{\UrlOrds}
\makeatother

\usepackage{booktabs}

\usepackage{xcolor}

\usepackage{listings}
\usepackage{pdfcomment}

\usepackage{amssymb}
\usepackage{pifont}
\ifCLASSOPTIONcompsoc
  \usepackage[%
    square,        
    comma,         
    numbers,       
    sort           
  ]{natbib}
\else
  \usepackage[%
    square,        
    comma,         
    numbers,       
    sort&compress 
  ]{natbib}
\fi

\usepackage{etoolbox}
\makeatletter
\patchcmd{\NAT@test}{\else \NAT@nm}{\else \NAT@hyper@{\NAT@nm}}{}{}
\makeatother

\usepackage[shortlabels]{enumitem}

\usepackage{multirow}

\usepackage{hyperref}
\hypersetup{hidelinks,
  colorlinks=true,
  allcolors=black,
  pdfstartview=Fit,
  breaklinks=true}
\usepackage[all]{hypcap}

\usepackage[capitalise,nameinlink]{cleveref}
\crefname{lstlisting}{\lstlistingname}{\lstlistingname}
\Crefname{lstlisting}{Listing}{Listings}

\ifcsmacro{minted}{}{%
  
}

\usepackage{xspace}

\DeclareFontFamily{U}{MnSymbolC}{}
\DeclareSymbolFont{MnSyC}{U}{MnSymbolC}{m}{n}
\DeclareFontShape{U}{MnSymbolC}{m}{n}{
  <-6>    MnSymbolC5
  <6-7>   MnSymbolC6
  <7-8>   MnSymbolC7
  <8-9>   MnSymbolC8
  <9-10>  MnSymbolC9
  <10-12> MnSymbolC10
  <12->   MnSymbolC12%
}{}
\DeclareMathSymbol{\powerset}{\mathord}{MnSyC}{180}

\ifCLASSOPTIONcompsoc
  \usepackage[caption=false,font=footnotesize,labelfont=sf,textfont=sf]{subfig}
\else
  \usepackage[caption=false,font=footnotesize]{subfig}
\fi

\usepackage{stfloats}

\usepackage{setspace} 
\usepackage{xspace}
\usepackage[super]{nth}

\usepackage[a4paper, total={184mm,239mm}]{geometry}

\begin{document}


\title{Leveraging Stochastic Depth Training for Adaptive Inference}

\author{%
 \IEEEauthorblockN{Guilherme Korol\IEEEauthorrefmark{1},
 Antonio Carlos Schneider Beck\IEEEauthorrefmark{2},
 Jeronimo Castrillon\IEEEauthorrefmark{1}\IEEEauthorrefmark{3}}
 \IEEEauthorblockA{\IEEEauthorrefmark{1} Chair for Compiler Construction, Dresden University of Technology, Dresden, Germany \\
    }
 \IEEEauthorblockA{\IEEEauthorrefmark{2}Institute of Informatics, Federal University of Rio Grande do Sul, Porto Alegre, Brazil 
  \\
 }
 \IEEEauthorblockA{\IEEEauthorrefmark{3}Center for Scalable Data Analytics and Artificial Intelligence (ScaDS.AI), Dresden, Germany 
  \\
 }
    \IEEEauthorrefmark{1}\{guilherme.korol,jeronimo.castrillon\}@tu-dresden.de,\IEEEauthorrefmark{2}caco@inf.ufrgs.br
}

\maketitle

\begin{abstract}
Dynamic DNN optimization techniques such as layer-skipping 
offer increased adaptability and efficiency gains but
can lead to i) a larger memory footprint as in decision gates, ii) increased training complexity (e.g., with non-differentiable operations), and iii) less control over performance-quality trade-offs due to its inherent input-dependent execution. 
To approach these issues, we propose a simpler yet effective alternative for adaptive inference with a zero-overhead, single-model, and time-predictable inference. Central to our approach is the observation that models trained with Stochastic Depth ---a method for faster training of residual networks--- become more resilient to arbitrary layer-skipping at inference time. We propose a method to first select near Pareto-optimal skipping configurations from a stochastically-trained model to adapt the inference at runtime later. Compared to original ResNets, our method shows improvements of up to 2$\times$ in power efficiency at accuracy drops as low as 0.71\%.
\end{abstract}


\section{Introduction}
\label{sec:intro}
Computing at the edge offers lower latency and reduced bandwidth usage by processing data close to the IoT devices, making it ideal for applications that demand fast responses. However, edge devices are constrained by limited computational power, memory, and energy, which can restrict their ability to handle more complex tasks, such as many resource intensive deep neural networks (DNN). As alternatives, numerous hardware and software optimizations have been proposed to reduce the computation cost of DNNs. 
Dynamic optimizations like layer skipping \cite{skipnet}
offer an intrinsically adaptive inference. Usually, this adaptation is performed on a per-sample basis. For example, in skipping, decisions gates are placed before every (set of) layer(s) and the decision of skipping (i.e., bypassing those layers) is taken by an additional set of dedicated layers of which computation does not contribute to the model result. This means that for every input, the DNN itself \emph{judges} how much computation is required for that particular input, accelerating the inference for \emph{easy} inputs, while \emph{hard} ones are processed more extensively~\cite{hapi,flexdnn,veit2018convolutional,jiang2020learning,DualDynamicInference}. 

Despite their success in adapting the inference processing, dynamic optimizations present one or more of the following drawbacks:
\textit{(a)} They increase the model's memory footprint 
due to decision gates. 
\textit{(b)} Many dynamic decisions, such as the skip-or-not ones, are based on non-differentiable operations that require methods like reinforcement learning for training, which has no convergence guarantees, complicates the training process \cite{bengio1994learning,elsken2019neural}, and can easily prevent deploying dynamic optimizations to new models.  \textit{(c)} The input-dependent behavior in dynamic optimizations makes those unfeasible to be used in, for example, real-time applications, where predictability is mandatory. These drawbacks hinder a wider adoption of adaptive inference for some application domains, thus limiting them from taking advantage of larger and better models.

\begin{figure}[!t]
\centering
\includegraphics[width=0.8\linewidth]{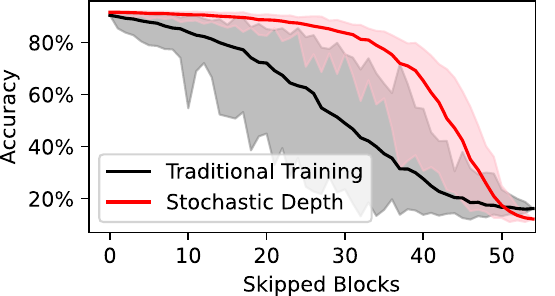}
\caption{Accuracy curve for skipping blocks in ResNet-110 (on CIFAR-10) trained with traditional and stochastic depth procedures (shaded area give the distribution for 500 skipping configurations).}
\label{fig:intro}
\end{figure}

In this work, we address the drawbacks above by \emph{keeping it simple}. More precisely, based on the idea of skipping (group of) layers, we propose a framework for inserting, at design time, our proposed \emph{simple and weightless} skip-or-not gates in front of the model's layers. The set of layers that will be skipped is also defined at design-time, and many skipping configurations with different combinations of layers to be skipped are generated. At runtime, these different skip configurations can be loaded into the skip-or-not gates. With that, we bring the heavily computational task of skipping decisions from runtime to design-time but still allowing for skipping adaptability after the model deployment. In other words, we enable an adaptive inference that is fully user-controllable (and thus, 
predictable) and does not need extra layers or any kind of learnable operators, not increasing 
memory footprint nor 
training complexity. As we will show, thanks to Stochastic Depth training \cite{stochastic} and a careful configuration of which layers to skip, carried by multiple near Pareto-optimal skipping configurations, our framework avoids major drops in accuracy that would be otherwise caused by arbitrarily skipping layers.

\autoref{fig:intro} gives an example of the accuracy improvements achieved by skipping layers (e.g., skipping a residual block) on a model trained with stochastic depth over skipping on one that was trained following a traditional training procedure. The figure shows the accuracy (y-axis) for a varying number of skipped blocks (x-axis) in a ResNet-110 \cite{resnet}. First, note the mean accuracy (black and red solid curves). The curves suggest a much greater resiliency to skipping for a model trained with stochastic depth -- precisely the behavior exploited in this work. Second, it is critical to note that there is a significant variability in the accuracy depending on which blocks are skipped (black and red shaded regions). For example, when skipping 37 blocks out of the 54 in ResNet-110, there is a 49.61\% accuracy gap between worst and best \emph{skipping configurations}. Therefore, it is important to select not only how many blocks to skip (which significantly impacts metrics like performance and energy) but also which blocks to skip to reduce the accuracy drop and enable different accuracy-performance inference profiles.

Concretely, this paper's contributions are as follows:
\begin{itemize}
    \item We propose a new gating mechanism that avoids the need of complex training methods and overhead-prone decisions layers. This mechanism enables, for the first time, repurposing Stochastic Depth training for deploying adaptive models;
    \item We propose a method for navigating the extensive skipping design space and deriving the Pareto front at design-time; and a runtime algorithm to adapt the inference processing on these configurations according to current edge requirements;
    \item We evaluate the proposed approach on a real edge platform against state-of-the-art adaptive techniques, showing improvements of up 2$\times$ in power efficiency and up to 1.97$\times$ in the number of processed inferences.
\end{itemize}

The state-of-the-art offers a plethora of DNN optimizations~\cite{compression_survey,laskaridis2021adaptive}. Here, we focus on the dynamic optimizations since they support adaptive inference techniques, which are especially important for deploying DNNs at the edge. 
Particularly, this work builds upon the skip optimization. The reasoning behind skipping is that not all samples require all of the network layers, which are necessary only for the hardest samples. For easy inputs, a shallow(er) network is \emph{enough}. The problem is usually treated as an optimization problem where the goal is to skip the most layers without dropping the network's accuracy. The prominent SkipNet~\cite{skipnet}, for instance, includes a method to \emph{train} decision gates that are placed before a group of layers. Based on the output of the previous layer, the gate can bypass the layers' computation. SkipNet features two types of decision gates: a feed-forward one (with one or two convolutional layers followed by a fully-connected one), and a recurrent gate (with a one-layer Long Short Term Memory, LSTM). Skipping is especially applicable to ResNets~\cite{resnet} due to their intrinsic design that already contains skips (residual connections) and their ensemble property.

\begin{figure}[!t]
\centering
\includegraphics[width=1\linewidth]{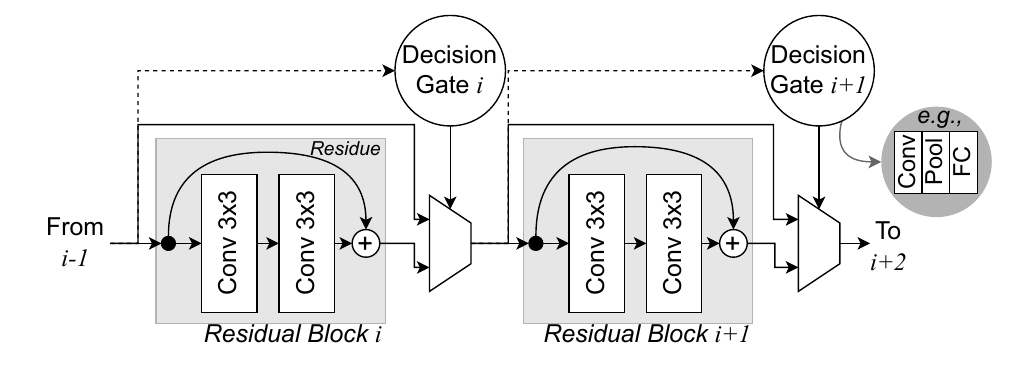}
\caption{Traditional skipping (adapted from \cite{skipnet}). Batch-normalization and activation layers omitted for simplicity.}
\label{fig:skipnet}
\end{figure}

\autoref{fig:skipnet} shows the traditional skipping approach (based on SkipNet). For each residual block (sequence of Conv 3$\times$3 with the respective ReLU and batch-normalization, omitted for simplicity), the Decision Gate reads in the previous feature map (dashed arrows) and selects whether to bypass the input feature map as the block's output (from the multiplexer's upper input) or use the feature map produced by the block (multiplexer's bottom input). Most of the literature on skipping networks differ on the configuration and training of the decision gate. A sample gate configuration is shown in \autoref{fig:skipnet}. The output of the Fully-Connected (FC) layer is discretized to control the bypass (e.g., bypass if greater than 0.5).

\subsection{Stochastic Depth}
\label{sec:stochDepth}

Stochastic Depth \cite{stochastic} is a popular training procedure that enables training very deep residual networks (e.g., ResNets with over 1200 layers). It emerged as a method to cope with challenges of training large networks, such as vanishing gradients and diminishing feature reuse~\cite{stochastic}. 
Key to stochastic depth is to have a \emph{deep} network for testing (delivering high accuracy) and a \emph{shallow} network during training (requiring less resources). 
More precisely, for each mini-batch, a set of layers is randomly disabled or dropped (i.e., for training on a shallower network). In the context of ResNets, this is done by passing forward the residue only (as in a identity function), thus not executing the layers that are part of the block, as will be further illustrated later.

Since initial layers extract \emph{low-level features} required by later layers, the authors propose that layers are dropped with a non-uniform probability. To this end, they use a probability function that decays with the depth, so that layers at the end of the network are more likely to be dropped. In a ResNet with $L$ blocks and hyper-parameter $p_L$, block $l$ is kept with a probability $p_l = 1 - \frac{l}{L}(1-p_L)$. For example, with a $p_L=0.2$, 40\% speedup in training is achieved for ResNet-110 on CIFAR-10 (with the same test error obtained with traditional training). In this work, however, our main interest in Stochastic Depth is not speeding-up training, but leveraging a property attributed to models trained with stochastic depth - a higher skipping resiliency. As shown in the plot of \autoref{fig:intro} and later in \autoref{sec:results}, models trained with a stochastic drop of layers become resilient to skipping after deployment.

\subsection{Related Work}

\begin{figure}[!t]
\centering
\includegraphics[width=1\linewidth]{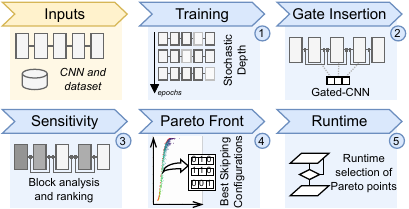}
\caption{Overview of the proposed framework.}
\label{fig:overview}
\end{figure}

ConvNet-AIG~\cite{veit2018convolutional} is a variation of skipping, in which layers are \emph{dedicated} to certain classes. This makes it possible to learn to skip layers related to not-interesting classes. To train the gates, authors model the gates as discrete random variables over two states (skip or not) and use the Gumbel-Max trick and its softmax relaxation. In the case of ConvNet-AIG, the skip decision can be interpreted as a decision based on the importance of a given layer to a given sample. Inspired by the biological brain, Jiang et al. proposed a two-branch network with skipping \cite{jiang2020learning}. Like the approach taken in SkipNet, the gates are trained \emph{with} the network weights. Interestingly, they report an improved robustness to adversarial attacks as byproduct of skipping. The Dual Dynamic Inference (DDI) framework~\cite{DualDynamicInference} uses skipping for IoT devices. It includes both layer and channel skipping and optional early-exits. Decision gates are trained with a supervised approach that employs a resource-aware loss that can be either in terms of number of floating-point operations (FLOPs) or energy. 

Besides skipping, optimizations like early-exit \cite{branchynet} also offer mechanisms for adaptive inference. Early-exit also works on a sample-by-sample basis by identifying the ``easier'' ones that can finish processing earlier by following exits (branching) connected to the main set of layers. For example, works like \cite{flexdnn}, \cite{hapi}, and \cite{classynet} introduce frameworks to optimize and deploy such models on embedded GPUs and edge environments. However, the required additional branches incur in a memory footprint overhead and, in case of not-taken exits, computation overhead. Additionally, by relying on a fully runtime data-dependent mechanism, early-exits (much like the traditional skipping) cannot offer a predictable inference time. 

Even though pruning \cite{channel_pruning}, another optimization, was initially proposed as a static optimization, it has also been extended to be used dynamically in the state-of-the-art. The pruning-created resource-accuracy trade-off was exploited on-device by many works like \cite{gromov2024unreasonable}, \cite{guo2024interpretable}, and \cite{adapex} (which even combines it with early-exit). Despite pruning returning a fixed model (with predictable runtime behavior), this optimization usually requires re-training the model to recover the lost accuracy. Retraining (or fine-tuning) the model, in turn, implies that multiple model versions need to be produced at design-time and kept at runtime if one wishes to adapt the inference processing, i.e., performing model switching.

\subsection{Our Contributions}
Our proposal contributes to the state-of-the-art on adaptive inference on the following. 
\textbf{\textit{Zero-Overhead}}: our simple yet effective gating mechanism avoids the need of complex training methods and avoids overhead-prone decisions layers (common on layer and block skipping models \cite{veit2018convolutional,jiang2020learning,DualDynamicInference}). 
\textbf{\textit{Predictability}}: for time-sensitive applications such as autonomous driving, it is imperative to deploy models that have predictable execution times. Our approach enables a fully controllable inference that is adaptable \textit{and} predictable (unattainable for input-dependent approaches \cite{early_exit,adapex}). 

\section{Adaptive Skipping with Stochastic Training}
\label{sec:overview}

\begin{figure}[!t]
\centering
\includegraphics[width=1\linewidth]{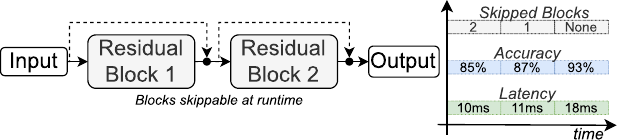}
\caption{A toy example of an adaptive ResNet with two skippable blocks.}
\label{fig:toy}
\end{figure}

\autoref{fig:overview} gives an overview of our framework for adaptive inference. 
It works in two phases. In the first phase, models are trained following the Stochastic Depth approach \cite{stochastic}, and the DNN design space is navigated to deliver operating points at the Pareto front for the skipping configurations (i.e., different set of skipped blocks deliver different accuracy-performance profiles). The generation of the Pareto front consists of four steps (1 to 4 in the figure). The second phase happens at runtime after the model has been trained and the Pareto front configurations have been selected. At runtime, the selected operating points (skipping configurations) can be switched at no cost (given this approach's single-model property, there is no need to load a different set of weights) to adapt the inference processing to current conditions. The list of configurations is passed to the runtime algorithm which, based on the profile of the current conditions at the edge, adapts the inference processing (step 5 in the \autoref{fig:overview}).

For instance, see the toy example on \autoref{fig:toy} with a ResNet with only two skippable blocks. There are three possible skipping configurations for adapting the inference processing, skipping block one, block two, or none. As explained earlier, these combinations will show different accuracy, latency, and energy profiles to be exploited dynamically in response to changes in runtime conditions or requirements at the edge. 

\subsection{Design-Time} 
In this section, we describe the main innovations regarding the design-time phase including the usage of Stochastic Depth for skipping-resilient models, the insertion of the fully-controllable gates, and the selection of skipping configurations for Pareto configurations.

\subsubsection{Inputs and Training}

The design-time phase starts by reading in the user CNN (in PyTorch) and the dataset. 
In Step 1 of this phase (see \autoref{fig:overview}), the CNN is trained with the Stochastic Depth procedure described in \autoref{sec:stochDepth}). 

\subsubsection{Gate Insertion}

After training, skip gates are inserted as PyTorch classes before the skippable layers (Step 2 in \autoref{fig:overview}). 
In this work, the Gated-CNN (gCNN, with the gates inserted)
is \emph{compiled} for deployment at the edge.
This means that we do not rely on high-level interpreted frameworks such as TensorFlow or PyTorch.
Instead, we leverage the Intermediate Representation Execution Environment (IREE)~\cite{iree}. 
IREE is an MLIR-based end-to-end compilation flow that accepts models described in multiple Deep Learning frameworks (e.g., PyTorch, Tensorflow, and ONNX). 
Central to the IREE compilation flow is the \texttt{torch-mlir} MLIR project for PyTorch and ONNX models.
For the gCNNs we extended \texttt{torch-mlir} to support lowering input-dependent gates as blocks controlled by \texttt{if} statements. 
This way, we seamlessly compile from Pytorch via ONNX onto edge platforms (Odroid XU4 boards, in our case study) using our modified IREE flow along with the IREE runtime (responsible for controlling the executable loading and its inputs and outputs). 
Apart from input images, a compiled gCNN also takes the skipping configuration in form of an array as input. 
Consequently, it is possible to change the model depth sample-by-sample, without needing to reload new weights from disk. 
That is, all skipping configurations lie within the \emph{same model and executable}.

\begin{figure}[!t]
\centering
\includegraphics[width=1\linewidth]{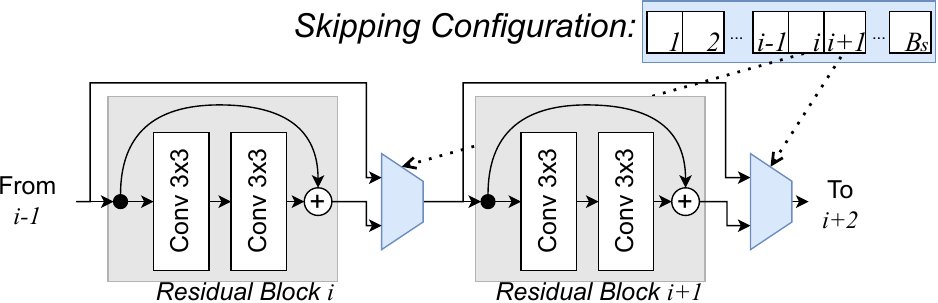}
\caption{Our proposed skipping. Batch-normalization and activation layers omitted for simplicity.}
\label{fig:our_skipnet}
\end{figure}

Recall that in ResNets there are segments (usually three or four) of residual blocks. All blocks of the same segment have the same number of channels, at every new segment the number of channels usually doubles (e.g., first segment with 16 channels, the next with 32, and the last one with 64 channels). 
In this work, all blocks other than the first in a segment are skippable and will have a gate placed in front. 
We define the size of the \textit{skipping configuration} as $B_s = B - Seg$, where $B$ is the number of blocks and $Seg$ is the number of segments. 
The skip configuration (illustrated in \autoref{fig:our_skipnet}) is an array $S = [s_1,s_2,..,s_{B_s}]$ where $s_i \in \{0,1\}$ defines whether the $i$-th block is skipped ($s_i=0$) or executed ($s_i=1$).

\subsubsection{Sensitivity Analysis}

A problem central to skipping is not only how much blocks should be skipped, but also which ones. Given the huge number of possible skipping configurations, evaluating all of them can quickly become untracktable. Formally, for selecting $N$ blocks among $B_s$ skippable blocks, there are ${B_s\choose N}$ options. Let us take the ResNet-110 as an example. This ResNet has 110 layers distributed in three segments with 18 blocks of two layers each (plus the input and last fully-connected layers). This give us $B_s = 54 - 3$ or $51$ skippable blocks. To select, let us say, $30$ blocks, there would be necessary $1.4\times10^{15}$ evaluations.

In this work, we approach this combinatorial problem by selecting which blocks to skip with a sensitivity analysis (Step 3 in \autoref{fig:overview})\footnote{In our experiments, ranking blocks with a sensitivity analysis performed better than other methods (i.e., l2-norm, fisher, hessian, and random search). Still, thanks to the modularity of the proposed framework, new methods can be easily incorporated.}.
The sensitivity analysis ranks blocks individually according to their accuracy impact (their \emph{importance}). Every block is removed (skipped) and the accuracy is evaluated. Then, an ordered list of blocks is created from the lowest accuracy (given by the removal of the most important block) to the highest accuracy (least important block). 

Regarding the time to build the sensitivity list, when compared to training the models, the list does not seriously increase training time. Precisely, it is necessary to run only one evaluation on the test set for every skippable block. As an example, take the training of ResNet-20 and -100. It took 500 epochs in our experiments, meaning 500 evaluations (plus the many more training iterations). In contrast, it takes only 7 evaluations (just 1\% of the evaluations run during training) and 51 evaluations (10\% of the training evaluations) for building the sensitivity lists for ResNet-20 and -100, respectively. If one considers also the time spent on the trainings set, the costs of building the sensitivity list decreases even further in comparison with the time for training those models.  

\subsubsection{Pareto Front Generation}

With the sensitivity list at hand, all filtered skipping configurations are evaluated on the complete test set according to their accuracy and execution time for deriving an approximation of the Pareto front (Step 4 in \autoref{fig:overview}). By ranging the number of skipped blocks $N$ from zero (meaning no skipping, $S = [1,1,..,1]$) to $N=B_s$ (where all blocks are skipped, $S = [0,0,..,0]$), a initial set of skipping configurations is generated. These configurations are obtained by skipping the least sensible $N$ blocks. Nevertheless, when evaluated, those configurations are not guaranteed to deliver Pareto operating points. That is, the sensitivity list regards blocks individually, only one is skipped at a time, and it does not evaluates their execution time. 

Therefore, in this step, the Pareto front is found by evaluating the accuracy \emph{versus} inference time of the skipping configurations. For example, out of the 51 possible skipping configurations in the evaluated ResNet-110, 32 configurations made into the Pareto front.
In the next step, the list of Pareto configurations supports the runtime adaptation (as a file, where rows, holding a Pareto $S$ configuration, are sorted by the number of skipped blocks).
\\

\subsection{Runtime}

\begin{algorithm}[!tb]
 \caption{Pseudocode for the runtime adaptation.}
 \footnotesize
 \label{alg:runtime}
 \begin{algorithmic}[1]
    \REQUIRE $Pareto$ with sorted configurations $cfg_i$ for $0<i<B_s$
    \REQUIRE $\Delta_{req}$ for maximum idle time before decreasing skipping
    \REQUIRE $min_{acc}$ for the minimum acceptable accuracy
    \STATE $Pareto \gets Pareto[i]\: \forall i \: (accuracy(cfg_i) > min_{acc})$
    \COMMENT{Filter out configurations with accuracy below the user's threshold}
    \STATE $curr\_cfg \gets Pareto[0]$
    \COMMENT{Start with no skipping}
    \LOOP
        \STATE $req=new\_request()$
        \COMMENT{Wait for new inference request}
        \IF{$busy()$}
            \STATE $drop\_inference(req)$
            \STATE $curr\_cfg \gets increase\_skipping()$
            \COMMENT{Increase number of skipped blocks by one if possible, otherwise returns $Pareto.last()$}
        \ELSE
            \IF{$t_{now} - t_{last} > \Delta_{req}$}
                \STATE $curr\_cfg \gets decrease\_skipping()$
                \COMMENT{Decrease the number of skipped blocks by one if possible, otherwise returns $Pareto.first()$}
            \ENDIF
            \STATE $process\_inference(req,curr\_cfg)$
            \STATE $t_{last} \gets t_{now}$
        \ENDIF
    \ENDLOOP
 \end{algorithmic} 
\end{algorithm}

On top of the IREE runtime, we developed an adaptive runtime to exploit the operating points at the Pareto front. It is in charge of selecting the skipping configuration that is fed to the executable along with the input image. The adaptive algorithm is presented in \autoref{alg:runtime}. 
First, the user's inputs ($\Delta_{req}$ and $min_{acc}$) along with the Pareto configurations are read. Then, all skipping configurations with accuracy below the minimum acceptable are discarded (line 1, done once).
After this initialization, at every new inference request (e.g., arrival of new input image in line 4), two outcomes are possible. 
If the device is busy, the inference requested cannot be processed and it is dropped (line 6).
The rationale behind it follows the MLPerf inference server standard \cite{mlperf} that defines a request as dropped if it arrives at a busy device. Therefore, in such cases, the runtime increases the number of skipped blocks by one (line 7). 
This way, the inference processing gradually moves to a faster skipping configurations in the Pareto Front. 
This adaptation mechanism only sacrifices accuracy when inference requests cannot be served due a lack of processing capabilities. 

As second outcome, when the device is free (else block in line 8),  the runtime can feed the executable with the input and current skipping configuration. 
However, if the time elapsed since the last request is greater than a pre-defined value ($\Delta_{req}$), the inference processing is assumed to safely go to a slower (but of higher accuracy) configuration (i.e., decrease the number of skipped blocks - line 10). 
In our experiments, we set $\Delta_{req}$ to the time taken by the zero-skipping model and $min_{acc}$ to 10\% below the accuracy of the original model. 
We also note that it is possible to incorporate more sophisticated methods in our solution to dynamically adapt the $\Delta_{req}$ and $min_{acc}$ parameters at runtime (e.g., with reinforcement learning methods).

\section{Methodology}
\label{sec:meth}

\subsection{CNNs and Datasets} 
We evaluate our approach on two datasets (CIFAR-10 \cite{cifar10} and CIFAR-100 \cite{cifar10}) and two residual models (ResNet-20 and ResNet-110). All input images follow CIFAR standard sizes of $32$x$32$. All models are trained for 500 epochs. Learning rate starts at $0.1$ for the first 250 epochs, $0.01$ for the next 175, and $10^{-4}$ until the end. All models are trained in PyTorch on an Nvidia Geforce RTX 3080. Accuracy reports are given on TOP-1 accuracy. After training, all models follow the same compilation steps, they are exported to ONNX files and compiled by our modified IREE flow with the same default compiler flags. Skipping configurations are saved as binary files and deployed with the executables.

We use two baselines for the evaluation, the \textbf{\textit{original}} ResNets without skipping and the state-of-the-art skipping approach, \textbf{\textit{SkipNet}}~\cite{skipnet}. 
Further, we follow SkipNet's supervised training of feedforward decision gates. For ResNet-20, the \texttt{ffgate1}~\cite{skipnet} gate is used consisting of an initial maxpool layer followed by two convolutional layers (both with batch normalization and ReLU), an average pool layer, and a final convolutional layer. For the larger ResNet-110, the evaluated SkipNet uses the \texttt{ffgate2} \cite{skipnet} gate configuration, which does not make use of an initial maxpool layer for dimension reduction (as in \texttt{ffgate1}). The decision to whether or not to skip in SkipNet is given by comparing the output neuron of the gates to a threshold $\alpha$. We set $\alpha=0.5$ following the default configuration in \cite{skipnet}.

\subsection{Evaluation Platform and Scenarios}
All evaluated models are deployed on Odroid XU4 board hosting an Exynos 5422 big.LITTLE chip with four Cortex-A15 and
four Cortex-A7 cores, fixed at frequencies of 1.8 GHz and 1.5 GHz respectively (inference processing is pinned at a big core). Performance is measured from within the IREE runtime that was modified to record inference time. Power and energy are measured using the ZES Zimmer LMG450 Power Analyzer connected to DC input with an external readout rate of 20Sa/s.

A workload generator requests 500 test images for evaluation (i.e., inference requests of batch size one). For ResNet-20 models, the workload generator sends one request every 5 seconds. For ResNet-110, it sends one request every 30 seconds. Besides, to emulate an edge environment with factors like Inference Per Second (IPS) fluctuation and network congestion, the base rate of inference requests (one every 5 or 30 seconds) varies over time~\cite{mlperf}, represented as 25\% random workload deviation every 10th request.

\section{Results}
\label{sec:results}

\subsection{Design Space}

\begin{figure*}[!t]
\centering
\includegraphics[width=1\linewidth]{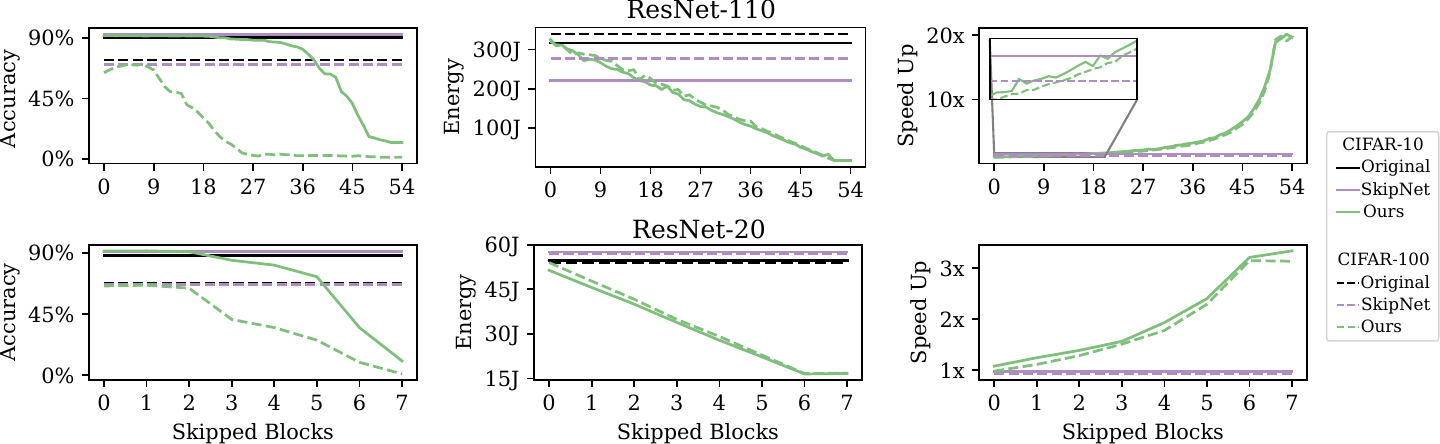}
\caption{Accuracy and Energy per infrence for original, SkipNet, and Ours on the left and center plots, respectively. Speedup over the Original baseline (right-most plots) for SkipNet and Ours. ResNet-110 on top plots, ResNet-20 on bottom plots. Dashed curves for CIFAR-100 dataset, solid curves for CIFAR-10. Note that original and SkipNet models are shown for comparison and do not follow the number of skipped blocks in the x-axis.
}
\label{fig:design-time}
\end{figure*}

\subsubsection{Accuracy}

\autoref{fig:design-time} presents the accuracy (left-side plots), energy consumption (center plots), and the speedup (right-side plots) for our approach over the number of skipped blocks (x-axis) for ResNet-110 (upper plots) and ResNet-20 (bottom plots) on CIFAR-10/100. Note that the baselines are shown for comparison only and are not following the number of skipped blocks in x-axis. 
From the accuracy plots, we see that our approach without skipping achieves a similar accuracy to the baselines. On CIFAR-10, the best accuracy is delivered by our approach at 91.44\% for ResNet-20 and by SkipNet at 92.67\% for ResNet-110. On CIFAR-100, the original models give the highest accuracy at 73.59\% and 67.37\% for ResNets-110 and -20, respectively. 
More importantly, the plot shows the adaptability opportunities enabled by our approach, by allowing full control over the model size (i.e., selecting the number of skipped blocks). For instance, for ResNet-110/CIFAR-10 (top-left plot, solid blue curve) the accuracy remains close (within 1\%) to the original one until around 20 blocks are skipped (out of the 54 possible). With 20 blocks, the accuracy drops only 0.23\% but 37\% of blocks are skipped. 
At a 10\% accuracy drop, 36 blocks can be skipped in the ResNet-110 on CIFAR-10 (delivering 81.51\%). Similar behavior is observed across the two evaluated datasets and ResNets. The gains from our skipping approach are especially clear for speedup and energy.

\subsubsection{Inference Time}
Despite Skipnet's high accuracy, we can see that the \textit{\textbf{overhead}} introduced by its decision gates can become an issue for inference time for some models and datasets. For instance, for ResNet-20, SkipNet ends up slowing down the inference when compared to the original model (speedups of $0.96\times$ and $0.92\times$ on CIFAR-10 and CIFAR-100, respectively). This is primarily due to SkipNet's low number of skipping blocks on ResNet-20 (from none to just three blocks are skipped). 
Therefore, the overhead from the layers in the decision gates become more apparent. Our approach, on the other side, presents significant speedup levels over the original ResNet-20 on CIFAR-10 (up to $3.33\times$) and CIFAR-100 (up to $3.13\times$). 

For the larger ResNet-110, SkipNet improves over the original model ($1.50\times$ speedup on CIFAR-10). For this model and dataset, SkipNet runs also faster than our approach for skipping configurations of up to 14 skipped blocks -- at which point SkipNet delivers an accuracy 1.18\% higher than our approach (\autoref{fig:design-time}). However, for configurations of over 14 skipped blocks, our approach delivers inferences faster than both baselines (up to $19.7\times$ over the original ResNet-110 on CIFAR-10). See the zoom-in from none to 20 skipped blocks in the top-center plot. 
Additionally, recall that the SkipNet curves in \autoref{fig:design-time} give the result over the full test set and due to its input-dependent behavior the inference time can vary drastically from input to input. On the evaluated models and datasets, we observe that SkipNet inference time can vary as much as 13$\times$ (between skipping all or executing all blocks), which severely impacts SkipNet's \textit{\textbf{predictability}}.

\subsubsection{Energy}
A behavior similar to inference time can be observed in energy consumption (center plots of \autoref{fig:design-time}). For instance, see the SkipNet's overhead on ResNet-20 w.r.t the original model ($3.46J$ overhead on CIFAR-10 and $2.93J$ on CIFAR-100). For the consumption of ResNet-110 (center top plot) on CIFAR-10, we can see that SkipNet saves energy over the original model (30\% energy savings). Regarding SkipNet against our approach for ResNet-110/CIFAR-10, SkipNet shows a more efficient processing for up to 15 skipped blocks. When more than 15 blocks are skipped, our approach shows the lowest energy consumption. For ResNet-110 at the larger dataset (CIFAR-100), our skipping starts to show energy consumption lower than SkipNet at around 10 blocks. It can be also observed that our approach delivers an almost linear improvement in energy consumption w.r.t the number of skipped layers - thanks to our \textit{zero-overhead gates}. In short, our approach makes the performance-energy-accuracy trade-off easily accessible (switching skipping configurations at no cost) by the user or runtime adaptive engines (as seen next).

\begin{table}[!t]
\centering
\caption{Accuracy, number of processed inferences, and inference per Watt for SkipNet and Our proposal w.r.t the original ResNet models. The higher, the better for all table.}
\label{tab:app}
\resizebox{\columnwidth}{!}{
\begin{tabular}{cccccccc}
\hline
\multicolumn{1}{l}{} & \multicolumn{1}{l}{} & \multicolumn{2}{c}{Accuracy} & \multicolumn{2}{c}{Proc. Inferences} & \multicolumn{2}{c}{Inf/Watts} \\
\multicolumn{1}{l}{} & \multicolumn{1}{l}{} & SkipNet & Ours & SkipNet & Ours & SkipNet & Ours \\ \hline
\multirow{2}{*}{CIFAR-10} & ResNet-20 & 2.83\% & -6.64\% & 0.94$\times$ & 1.48$\times$ & 0.97$\times$ & 1.54$\times$ \\
 & ResNet-110 & 2.27\% & -3.13\% & 1.47$\times$ & 1.97$\times$ & 1.46$\times$ & 2.00$\times$ \\
\multirow{2}{*}{CIFAR-100} & ResNet-20 & 0.26\% & -2.40\% & 0.94$\times$ & 1.15$\times$ & 0.95$\times$ & 1.18$\times$ \\
 & ResNet-110 & 2.88\% & -0.71\% & 1.28$\times$ & 1.22$\times$ & 1.29$\times$ & 1.26$\times$ \\ \hline
\end{tabular}
}
\end{table}

\subsection{Runtime Evaluation}

\autoref{tab:app} summarizes the runtime evaluation (recall from \autoref{sec:meth}) on the average accuracy, total number of processed inferences, and the processed inferences per Watt ratio. All results are presented w.r.t the Original baseline for the sake of simplicity. First, note the clearly higher accuracy delivered by SkipNet for most models and datasets. The exception is ResNet-20 on CIFAR-100, in which the original model delivered an accuracy 0.26\% higher than that of SkipNet. 
Throughout our evaluation, the accuracy drops of our approach lie within the 10\%, defined by the parameter $acc_{min}$ (see \autoref{alg:runtime}). 
This leads to great improvement in inferences and inferences per Watt, as shown in the table.

In terms of the number of processed inferences, our approach achieves higher performance than both baselines in almost all evaluation cases. 
The best case scenario for our adaptive inference happens at the ResNet-110 on CIFAR-10, where it almost doubled (1.97$\times$) the number of processed inferences (at a small 3.13\% accuracy drop) when compared to the original baseline. The exception is the ResNet-110/CIFAR-100 evaluation, where SkipNet achieved the highest performance. This is due to the steeper accuracy drop with our skipping approach (recall the accuracy plot in \autoref{fig:design-time}). With less operating points (i.e., skipping configurations) delivering accuracy above $acc_{min}$, our runtime has less operating points available to adapt the inference and ends up with a speedup over the the other two baselines smaller than in the other evaluation cases.

Another important factor for edge deployment is the power efficiency (see right-most columns of \autoref{tab:app}).
The last column highlights the power efficiency delivered by our zero-overhead, controllable skipping. 
Even though SkipNet achieves the best results for the ResNet-110 on CIFAR-100, in all the other evaluation cases, our gated models achieve better efficiency and, contrarily to SkipNet, is constantly better than the original ResNet-20 and ResNet-100 models.
original ResNet-20 and ResNet-100 models.

\section{Conclusions}
\label{sec:conc}

Based on the observation that CNNs become more resilient to arbitrary skipping when trained with the stochastic depth, our framework trains CNNs that can be leveraged for fully-controllable, zero-overhead, runtime adaptive inference as gated CNNs. However, selecting which layers to skip (i.e., the skipping design space) is a combinatorial problem which quickly becomes intractable. 
To cope with this, we compute a sensitivity list of the model's layers to approximate the Pareto front of skipping configurations. At runtime, the generated operating points are exploited to adapt the inference processing.

\section*{Acknowledgments}
This work was funded in part by the AI competence center ScaDS.AI Dresden/Leipzig in Germany (01IS18026A-D) and by the EU Horizon Europe Programme under grant agreement No 101135183 (MYRTUS). Views and opinions expressed are however those of the author(s) only and do not necessarily reflect those of the European Union. Neither the European Union nor the granting authority can be held responsible for them.



\bibliographystyle{IEEEtranN} 
\bibliography{paper}

\end{document}